\documentclass[sigconf]{acmart}

\usepackage{multicol}
\usepackage{multirow}
\usepackage{graphicx}
\usepackage{subfigure}
\usepackage{makecell}
\usepackage{amsmath,mathtools,amsthm}
\usepackage{color}
\usepackage{algorithm}
\usepackage{algorithmic}
\usepackage{bbold}
\usepackage{colortbl}

\AtBeginDocument{%
  \providecommand\BibTeX{{%
    \normalfont B\kern-0.5em{\scshape i\kern-0.25em b}\kern-0.8em\TeX}}}

\copyrightyear{2023}
\acmYear{2023}
\setcopyright{acmlicensed}
\acmConference[MM '23] {Proceedings of the 31st ACM International Conference on Multimedia}{October 29--November 3, 2023}{Ottawa, ON, Canada.}
\acmBooktitle{Proceedings of the 31st ACM International Conference on Multimedia (MM '23), October 29--November 3, 2023, Ottawa, ON, Canada}
\acmPrice{15.00}
\acmISBN{979-8-4007-0108-5/23/10}
\acmDOI{10.1145/3581783.3612312}
\begin{document}
\title{Echoes: Unsupervised Debiasing via Pseudo-bias Labeling in an Echo Chamber}

\author{Rui Hu}
\affiliation{%
  \institution{Beijing Jiaotong University, China}
  \state{}
  \country{}
}
\email{rui.hu@bjtu.edu.cn}

\author{Yahan Tu}
\affiliation{%
  \institution{China University of Geoscience Beijing, China}
  \state{}
  \country{}
}
\email{yahan.tu@cugb.edu.cn}

\author{Jitao Sang}
\affiliation{%
  \institution{$^1$Beijing Jiaotong University, China}
  \institution{$^2$Peng Cheng Lab, China}
  \state{}
  \country{}
}
\email{jtsang@bjtu.edu.cn}



\begin{abstract}
\urlstyle{tt}
Neural networks often learn spurious correlations when exposed to biased training data, leading to poor performance on out-of-distribution data. A biased dataset can be divided, according to biased features, into bias-aligned samples (i.e., with biased features) and bias-conflicting samples (i.e., without biased features). Recent debiasing works typically assume that no bias label is available during the training phase, as obtaining such information is challenging and labor-intensive. Following this unsupervised assumption, existing methods usually train two models: a biased model specialized to learn biased features and a target model that uses information from the biased model for debiasing. This paper first presents experimental analyses revealing that the existing biased models overfit to bias-conflicting samples in the training data, which negatively impacts the debiasing performance of the target models. 
To address this issue, we propose a straightforward and effective method called \textit{Echoes}, which trains a biased model and a target model with a different strategy. We construct an "echo chamber" environment by reducing the weights of samples which are misclassified by the biased model, to ensure the biased model fully learns the biased features without overfitting to the bias-conflicting samples. The biased model then assigns lower weights on the bias-conflicting samples. Subsequently, we use the inverse of the sample weights of the biased model for training the target model. Experiments show that our approach achieves superior debiasing results compared to the existing baselines on both synthetic and real-world datasets. Our code is available at \textcolor{magenta}{\url{https://github.com/isruihu/Echoes}}.

\end{abstract}





\maketitle

\section{Introduction}

\begin{figure}
\setlength{\abovecaptionskip}{1em} 
\setlength{\belowcaptionskip}{-1em} 
    \centering
    \includegraphics[width=1\linewidth]{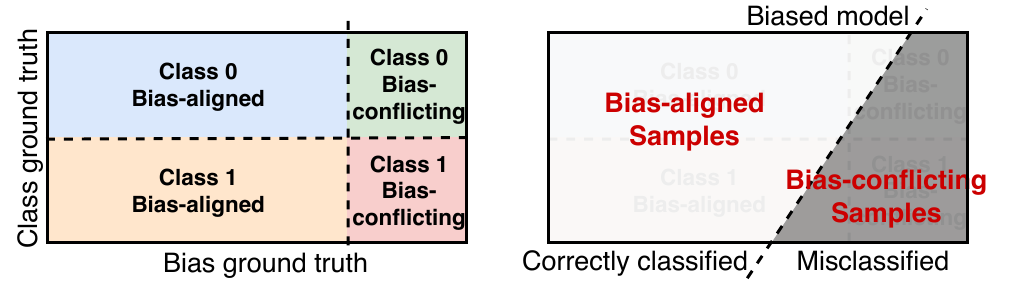}
    \caption{Illustration of using a biased model to differentiate  the training samples in terms of bias. Left is the ground truth labels of data samples and right is the division result (i.e., pseudo-bias labels) by the predictions of the biased model.}
    \label{fig:pseudo-bias-label}
\end{figure}

Despite the success of deep neural networks (DNNs), recent studies have revealed that many models have a tendency to learn shortcut cues~\cite{geirhos2020shortcut, li_2023_whac_a_mole}. DNNs often pick up simple, discriminative cues from the training data, and ignore more complex but important ones~\cite{shah2020pitfalls, teney2022evading}. The correlations between the simple cues and the labels are often spurious. A DNN that relies on such simple spurious correlations for predictions will have poor performance on out-of-distribution (OOD) data, i.e., data without such correlations~\cite{hendrycks2021many,tahir2022distributional}. The datasets with spurious correlations are typically said to be \textit{biased}. Models trained on biased datasets that contain spurious correlations of protected attributes (e.g., gender, race) and labels may suffer from fairness problems.
Biased datasets typically contain a majority of \textit{bias-aligned} samples and a minority of \textit{bias-conflicting} ones. For bias-aligned samples, ground truth labels are correlated with both target features and biased features. While for bias-conflicting samples, labels are correlated only with target features. As an example, suppose that the bird bodies and the backgrounds are target features and biased features for bird classification, respectively, and as birds often co-occur with the sky, birds in the sky are \textit{bias-aligned}, and birds on the land are \textit{bias-conflicting}.
Clearly, maximizing the effect of bias-conflicting samples is the key point to preventing models from learning spurious correlations.

 Existing debiasing methods can be divided into \textit{supervised} and \textit{unsupervised} debiasing in terms of the availability of bias information in the training phase. For supervised debiasing, some methods require bias annotations, i.e., each training sample is provided with a bias label~\cite{Sagawa2020Distributionally, tartaglione2021end, hong2021unbiased}. Other methods use prior knowledge of the bias type, which enables them to design a debiasing network tailored for the predefined bias type~\cite{bahng2020learning, GeirhosRMBWB19}. There are limitations to supervised debiasing: bias mitigation cannot be performed when bias labels or types are not available, and it requires expensive labor to manually identify the bias types and get bias labels. Since accessing bias information in the training phase is unreasonable, recent works focus on the more challenging and realistic task - unsupervised debiasing, which mitigates the bias without annotation. 

Unsupervised debiasing methods work with the assumption that biased features are easier to be learned than target features, meaning that they are picked up earlier by the model during training~\cite{nam2020learning, liu2021just, lee2021learning}. Following this assumption, the works typically train two models: one is an auxiliary biased model specialized to learn biased features so that the predictions of the biased model can be used as pseudo-bias labels, i.e., whether a sample is bias-aligned or bias-conflicting (See Fig.\ref{fig:pseudo-bias-label} for illustration). And the other is the desired target model that using pseudo-bias labels from the biased model for debiasing. Therefore, a well-designed biased model is critical to improve the debiasing performance of the target model~\cite{lee2022biasensemble}.

However, our experimental analysis reveals that existing bias models, e.g., an ERM model~\cite{liu2021just} or a GCE model~\cite{nam2020learning, lee2021learning} (see Sec~\ref{sec:2.1}), fail to provide accurate pseudo-bias labels, i.e., unable to identify bias-conflicting samples in the training data. Due to the scarcity of the bias-conflicting samples, the biased models are prone to memorize these samples to minimize the training loss, leading to the overfitting on them. For example, the error rate of an ERM model on the bias-conflicting sample of CelebA~\cite{liu2015faceattributes} is close to zero at the end of training. This, in turn, negatively impacts the debiasing performance of the target modes. Besides, through experiments, we find that partially biased data is sufficient for the model to learn biased features as they are easy-to-learned.

To address this, we propose a simple and effective unsupervised debiasing method called \textbf{Echoes} (\textbf{E}cho \textbf{ch}amber f\textbf{o}r d\textbf{e}bia\textbf{s}ing) which trains a biased model and a target model in an echo chamber. 
Unlike the training of existing biased model, we are inspired by the \textit{Echo chamber}~\footnote{In media, an echo chamber is an environment or ecosystem in which participants encounter beliefs that amplify or reinforce their preexisting beliefs by communication and repetition inside a closed system and insulated from rebuttal.} phenomenon in media communication, and continually adjust the weights of the samples during the training phase like the information changing in an echo chamber, so that the biased model can distinguish bias-conflicting samples from bias-aligned ones. Specifically, we consider that in the early training phase, the majority of samples correctly classified by the biased model are bias-aligned, while the samples misclassified by the model consist of a minority of bias-aligned samples and a majority of bias-conflicting ones. Then we can "hide" these error samples by reducing their weights. The remaining samples that are correctly classified by the biased model will be retrained in the next training round, just like the echo in an echo chamber. In the end, the biased model sufficiently learns from bias-aligned samples, while underfitting for bias-conflicting ones, which indicates that more accurate bias information can be provided to the target model.

\begin{figure}
\setlength{\abovecaptionskip}{0em} 
\setlength{\belowcaptionskip}{-1em} 
    \centering
    \includegraphics[width=0.71\linewidth]{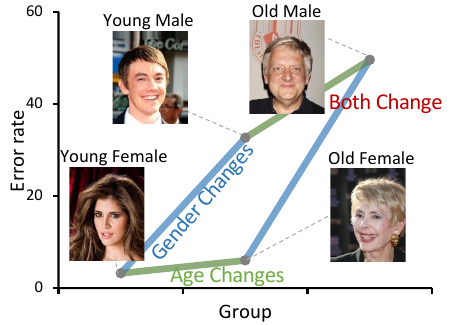}
    \caption{Illustration of multi-bias learning in \textit{Smiling} classification. Model error rates increases when test samples change from bias-aligned (e.g., young female) to single bias-conflicting (e.g., young male), and reaching a maximum when both bias conflict (e.g., old male).}
    \label{fig:multi-bias-instruction}
    
\end{figure}

In the echo chamber, a lower sample weight indicates that it is more often misclassified by the biased model, and a higher weight indicates that it is more often correctly classified. This means that samples with lower weights are harder to be learned by the biased model, making them more important for the target model. Based on this analysis, we simply use the inverse of the biased model's sample weights as the target model's sample weights, so that the target model can focus on learning hard samples to prevent learning spurious correlations.

We evaluate our method on both the \textit{single-biased} benchmarks that are commonly used in existing works and the \textit{multi-biased} benchmarks that are more challenging and rarely considered. Fig.\ref{fig:multi-bias-instruction} illustrates the phenomenon of multi-bias learning. Experimental results suggest that our method consistently outperforms prior methods in both multi-bias and single-bias settings. Ablation studies show the advantages of our biased model in biased feature learning compare to existing ones.

In summary, our contributions are as follows:
\begin{itemize}
    \item We find that existing biased models for unsupervised debiasing fail to provide accurate pseudo-bias labels, which negatively impacts the debiasing performance.
    \item We propose a simple unsupervised debiasing method called Echoes, which trains an auxiliary biased model and a target model in an echo chamber.
    \item We perform extensive experiments with single-biased and multi-biased benchmarks, demonstrating that Echoes outperforms the existing methods.
\end{itemize}

\section{A close look at the training of biased models}

This section describes the details of two analytical experiments in which we observe the overfitting of the biased model on the training data, and partially biased data is sufficient for the model to learn biased features. In Section~\ref{sec:2.1}, we first introduce the common choices of biased models, that we utilize for the observation. Then, we elaborate the results of the experiments in Section~\ref{sec:2.2} and ~\ref{sec:2.3}.

\subsection{Background}
\label{sec:2.1}
Since annotating bias labels or identifying the bias types in advance is challenging and labor intensive, recent works consider an auxiliary model that specializes in learning biases to guide the training of the debiased model~\cite{liu2021just, nam2020learning, lee2021learning}. A good biased model should sufficiently learn biased features to have the ability to differentiate the training samples in terms of the biases. For instance, correctly classifying bias-aligned samples and misclassifying bias-conflicting ones, so that the target model can focus on learning the wrong samples based on the predictions of the biased model.

\begin{figure}[!th]
\setlength{\abovecaptionskip}{0em} 
\setlength{\belowcaptionskip}{-1em} 
    \centering
    \subfigure[CelebA]{
        \includegraphics[height=80px]{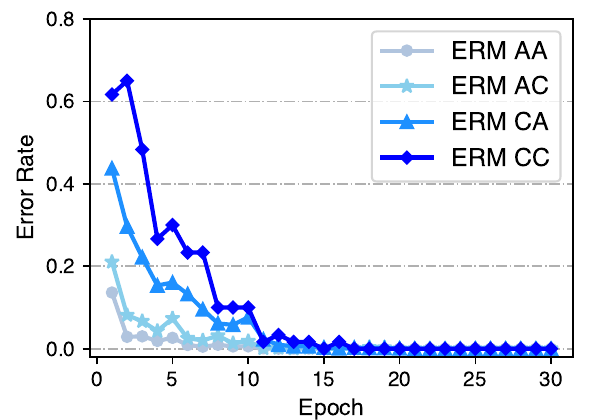}
        \includegraphics[height=80px]{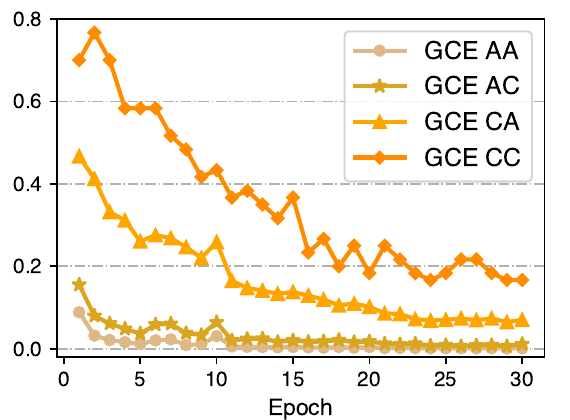}
    }
    \subfigure[UrbanCars]{
        \includegraphics[height=80px]{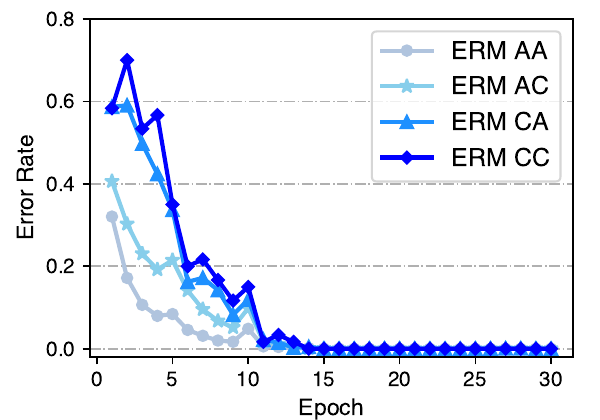}
        \includegraphics[height=80px]{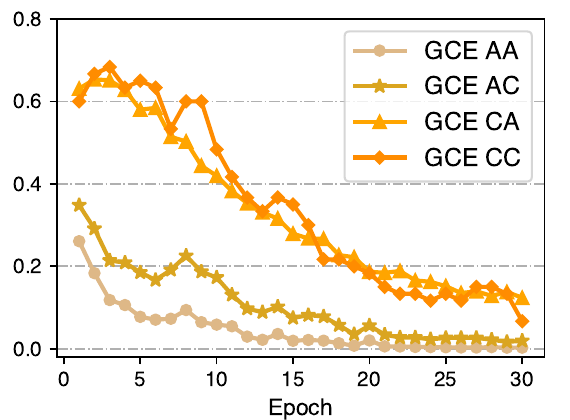}
    }
    \caption{The classiﬁcation error rates of ERM and GCE models on the training data of CelebA and UrbanCars.}
    \label{fig:error-rate}
\end{figure}

There are two common choices for a biased model: an ERM model or a GCE model. Liu et al.~\cite{liu2021just} train an ERM model as an identification model to select samples with high training loss. ERM minimizes the following objective:
\begin{equation}
    J_{ERM}(\theta) = \frac{1}{n}\sum_{i=1}^{n} \ell (x_i,y_i;\theta)
\end{equation}

\noindent where $\ell(x_i,y_i;\theta)$ is the cross-entropy loss of the model parameterized by $\theta$ for the sample $(x_i,y_i)$. They consider that ERM models tend to fit samples with easy-to-learned spurious correlations, i.e., bias-aligned samples. Then they train a final model by upweighting the points in the error set identified by the ERM model. In addition, Nam et al.~\cite{nam2020learning} use the Generalized cross entropy (GCE) loss~\cite{DBLP:conf/nips/ZhangS18} to encourage the model to focus on the samples with high probability values so that amplify the bias. The GCE loss is defined as:
\begin{equation}
    GCE(p(x;\theta), y) = \frac{1-p_y(x;\theta)^q}{q}
\end{equation}
where $p(x;\theta)$ and $p_y(x;\theta)$ are softmax outputs of the model and the softmax probability of the target class $y$, respectively, and $q \in (0,1]$ is a hyperparameter that controls the degree of amplification. In a biased dataset, the samples for which the model has high probability values are mostly bias-aligned samples, the GCE loss encourages the model to focus on such samples, leading to the model to be biased.

\subsection{Biased Models Overfit to the Bias-conflicting Samples}
\label{sec:2.2}
To confirm whether the existing biased models can achieve the expectation, i.e., discriminating the training samples in terms of the biases, we train ERM model and GCE model on CelebA~\cite{liu2015faceattributes} and UrbanCars~\cite{li_2023_whac_a_mole}. Both datasets have one target attribute and two biased attributes. In CelebA, the target attribute is \textit{smiling} and the biased attributes are \textit{gender} and \textit{age}. In UrbanCars, the target attribute is \textit{car} and the biased attributes are \textit{background} and \textit{co-occurring object} (see Sec~\ref{sec:4.1} for datasets details). 
\begin{table}[!th]
    \caption{The bias-aligned accuracy on the test sets of CelebA and UrbanCars. Training dataset size is 1.0 means using 100\% training data for model training.}
    \centering
    \begin{tabular}{ccc}
    \toprule
        \multirow{2}{*}{Train dataset size} & \multicolumn{2}{c}{Bias-aligned Acc.} \\ \cline{2-3}
                                            & CelebA & UrbanCars \\ \hline
        1.0 & 98.4 & 92.4 \\
        0.5 & 99.2\footnotesize{(\textcolor{orange}{+1.0\%})} & 86.0\footnotesize{(\textcolor{blue}{-6.9\%})} \\
        0.2 & 94.4\footnotesize{(\textcolor{blue}{-4.1\%})} & 85.2\footnotesize{(\textcolor{blue}{-7.8\%})} \\
        0.1 & 94.0\footnotesize{(\textcolor{blue}{-4.5\%})} & 80.0\footnotesize{(\textcolor{blue}{-13.4\%})} \\
    \bottomrule
    \end{tabular}
    \label{tab:partially-biased-data}
    \vspace{-0.7cm}
\end{table}
Then, the training data can be divided into four groups based on the different biased attribute values. For example, the CelebA dataset can be divided into
$\{$young female, old female, young male, old male$\}$, where the first group is the bias-aligned samples and the other three groups are the bias-conflicting samples. 
For convenience, we use \textbf{AA} to refer to the group which is aligned on both biased attributes (e.g., young female), \textbf{AC} to the group which is aligned on the first biased attribute and conflicting on the second biased attribute (e.g., old female), \textbf{CA} to the group which is conflicting on the first biased attribute and aligned on the second biased attribute (e.g., young male), and \textbf{CC} to the group which is conflicting on both biased attributes (e,g., old male), respectively. 

~\\
\noindent\textbf{Overfitting on the bias-conflicting samples.} We train the models for 30 epochs, and after each epoch, we calculate the error rate of the biased models on each group of training data. Fig.\ref{fig:error-rate} shows the experimental results. Apparently, as the training progresses, the models overfit to the bias-conflicting samples (AC, CA and CC). The error rates of all groups of the training data decrease, indicating that the biased model becomes less capable of discriminating between samples in terms of bias. Such performance is not consistent with what we expect from the bias models.

~\\
\noindent\textbf{Early stopping is not good enough.} Early stopping~\cite{yao2007early} is a common strategy to prevent overfitting of the model. However, for the training of biased models, there are several limitations to applying early stopping. On the one hand, early stopping usually requires a validation set for deciding when to stop. However, in our setting, we do not have information on bias and therefore cannot construct a validation set that can be used to prevent the model from overfitting to bias-conflicting samples. On the other hand, one can select models that are in the very early stages of training (e.g., models trained in one or two epochs). The models have not been sufficiently trained at that time and therefore not yet overfit to the bias-conflicting samples. However, from Fig.\ref{fig:error-rate}, at the beginning of training, the biased features are also not fully learned by the models, leading to a non-negligible error rate on the bias-aligned samples. And due to the large number of bias-aligned samples, even a small error rate can lead to a number of bias-aligned samples being categorized by the model as bias-conflicting samples, which is detrimental to the training of the target model.

\subsection{Part of Biased Data is Sufficient for Bias Learning}
\label{sec:2.3}

\begin{figure}[!thb]
\setlength{\abovecaptionskip}{0em} 
\setlength{\belowcaptionskip}{-1em} 
    \centering
    \includegraphics[width=1\linewidth]{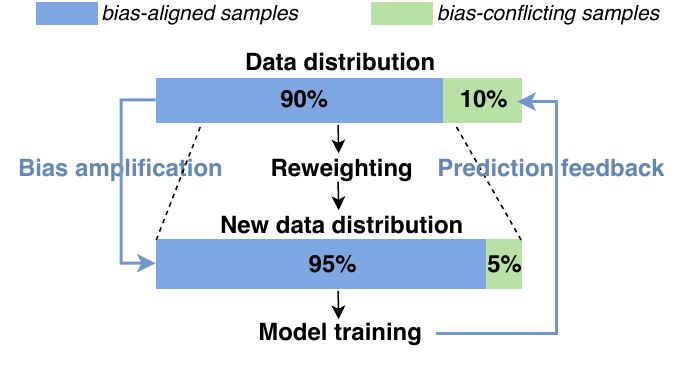}
    \caption{The training process of the biased model. The data distribution becomes increasingly biased with training, just like an echo chamber.}
    \label{fig:model-biased}
\end{figure}

In addition to early stopping, reducing the size of the training data can mitigate model overfitting because the model cannot overfit to unseen samples. And we find part of biased data is sufficient for the biased model to learn biased features.

We train the ERM models on different ratios of the CelebA~\cite{liu2015faceattributes} and UrbanCars~\cite{li_2023_whac_a_mole} datasets. To be specific, we use 100\%, 50\%, 20\%, and 10\% of the training samples to train the model respectively. For each ratio, we train the model for 100 epochs, and report accuracy of bias-aligned samples of test data, which reflects how well the model learned for biased features. Table~\ref{tab:partially-biased-data} shows the results. The accuracy of the model on bias-aligned samples mostly decreases when the size of training data decreases. But the decrease in accuracy is small compared to the decrease in data size. For example, training with only 20\% of the UbanCars training samples, the model accuracy dropped by only less than 8\% compared to training with 100\% of samples.


\section{Method}

Motivated by such an observation in Section~\ref{sec:2.3}, we first introduce our training strategy of a biased model, which assumes that biased features are "easy-to-learned" than target ones (Sec.\ref{sec:3.1}). Then, we introduce a debiasing approach based on samples re-weighting (Sec.\ref{sec:3.2}). Finally, we describe the complete method that combines the debiasing approach with our auxiliary biased model (Sec.\ref{sec:3.3}).

\subsection{Biased Model}
\label{sec:3.1}

\noindent\textbf{Training goal.} In a highly biased dataset, bias-conflicting samples constitute only a small fraction of the training data, we must maximize the effect of them to prevent the target model
from learning spurious correlations. However, through the observation in Section~\ref{sec:2.2}, both ERM and GCE models overfit the bias-conflicting samples in the training data. The goal of training a biased model is thus to ensure that the bias model fully learns the biased features while protecting the bias-conflicting sample from being learned.

\begin{algorithm}[!thb]
    \caption{Training of the biased model}
    \label{alg:biased-model}
    \begin{algorithmic}[1] 
        \STATE \textbf{Input}: dataset $D=\{(x_i,y_i)\}$, model $f_\theta$, number of iterations $T$, batch size $B$, sample weights $W_B$, hyperparameter $\alpha$.
        \STATE \textbf{Output}: a biased model $f_\theta$
        \STATE Initialize $\theta$, and initialize $W_B=\mathbb{1}$
        \STATE \textbf{for} t = 1,\dots,T \textbf{do}
        
        \STATE \qquad \textit{/** Train model $f_\theta$ **/}
        \STATE \qquad \textbf{for} b = 1,\dots,B \textbf{do}
        \STATE \qquad \qquad Draw a batch $\mathcal{B}=\{(x_i, y_i)\}_{i=1}^{n}$ 
        \STATE \qquad \qquad $\theta \leftarrow \theta - \nabla W\mathcal{L}_{CE}(\mathcal{B})$
        \STATE \qquad \textbf{end for} \item[]

        \STATE \qquad \textit{/** Update weights $W$ **/}
        \STATE \qquad Get predictions $\hat{Y}=\{\hat{y}_i\}$ of the model $f_\theta$ on $D$
        \STATE \qquad Count the error rate for each class $E=\{e_i\}_{i=1}^C$
        \STATE \qquad \textbf{for} $c=1,\dots,C$, \textbf{do}
        \STATE \qquad \qquad \textbf{if} $e_c<t_{error}$ \textbf{do}
        \STATE \qquad \qquad \qquad $W_c \leftarrow \alpha W_c$
        \STATE \qquad \qquad \textbf{end if}
        \STATE \qquad \textbf{end for}
        \STATE \textbf{end for}
    \end{algorithmic}
\end{algorithm}

~\\
\noindent\textbf{Constructing an echo chamber.} To achieve the goal, our idea is to constructing an "echo chamber" environment when training the biased model. Based on the assumption that the model gives priority to learning biased information, we assign weights in the echo chamber to achieve homogeneity and diffusion of information during sample selection, thus achieving full learning of biased features. Our insight is that samples that are correctly classified are easy-to-learned for the biased model, i.e., mostly bias-aligned samples, while samples that are misclassified by the model contain a majority of bias-conflicting samples and a minority of bias-aligned samples. Therefore, reducing the weights of misclassified samples is equivalent to constructing an echo chamber environment that allows the model to learn only the samples it tends to learn.

~\\
\noindent\textbf{Approach.} Fig.\ref{fig:model-biased} illustrates the approach. At the beginning of the training, we have a biased dataset that each sample has with equal weights for each sample (e.g., 1). Then, after the model is trained for one epoch, each sample is re-weighted with the model's prediction results. The re-weighting rule can be formulated as:
\begin{equation}
    w^i(x) = \left\{ 
        \begin{array}{ll}
            w^{i-1}(x), & if \; f_{\theta}(x) = y \\
            \alpha w^{i-1}(x), & if \; f_{\theta}(x) \neq y
        \end{array}
    \right.
\end{equation}
where $w^i(x)$ is the weight of sample $x$ at the $i$-th epoch, $f_\theta(x)$ is the prediction result of the biased model for the sample $x$, $y$ is the ground truth label, and $\alpha \in [0, 1] $ is a hyperparamter that controls the degree of weight reduction of the sample. If $\alpha=0$, it is equivalent to removing the samples that the model predicts incorrectly from the training data after each epoch, and if $\alpha=1$, it degenerates into standard ERM training. To improve the stability of the training, we use a hyperparameter $t_{error}$ (i.e., the threshold of error rate) to control whether to perform the re-weighting operation after a training epoch. We only re-weight those samples which are in a class with an error rate less than $t_{error}$. The high error rate of a class indicates that the model has not learned the features of this class well, thus the error samples of this class are not bias-conflicting at this time. We define the biased loss as follows:
\begin{equation}
    L_{biased} = W_{biased}CE(P(X;\theta), Y)
\end{equation}
where $CE$ is the cross-entropy loss and $P(X;\theta)$ is the softmax output of the biased model. The pseudo-code of the training of the biased model is presented in Alg.~\ref{alg:biased-model}.

\begin{figure*}[!thb]
    \centering
    \includegraphics[width=.8\linewidth]{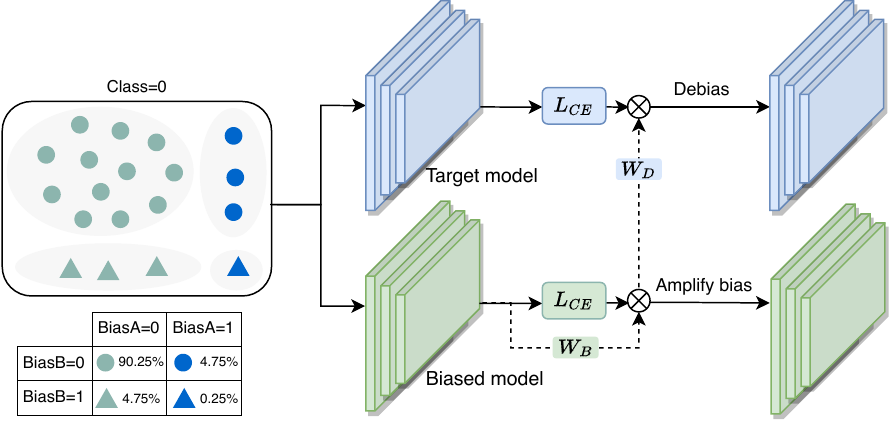}
    \caption{Illustration of our proposed method \textit{Echoes}. We jointly train a biased model and a target model with the biased training data, where the sample weights $W_B$ (for the biased model) and $W_D$ (for the target model) change continuously as the models training.}
    \label{fig:model}
\end{figure*}

\subsection{Target Model}
\label{sec:3.2}

\noindent\textbf{Training goal.} The training goal of the target model $f_D$ is to learn debiased representation. Since we do not have bias information about the training data, we need to use the information from the biased model $f_B$ to guide the training of the target model.

~\\
\noindent\textbf{Debiasing $f_D$ via $f_B$.} Since biased features are "easy-to-learned" than target ones, the cross-entropy loss values of bias-conﬂicting samples are relatively higher than those of bias-aligned ones. Recent works~\cite{nam2020learning, seo2022unsupervised} then re-weight the sample weights $w(x)$ for training the target model with such a characteristic. Speciﬁcally, Nam et al.~\cite{nam2020learning} formulated $w(x)$ as:
\begin{equation}
    w(x) = \frac{\mathcal{L}_{CE}(f_B(x),y)}{\mathcal{L}_{CE}(f_B(x),y) + \mathcal{L}_{CE}(f_D(x),y)}
\end{equation}
where $f_B(x)$ and $f_D(x)$ indicate the prediction outputs of $f_B$ and $f_D$, respectively. With this formula, the bias-conflicting samples will have high re-weighting values, and bias-aligned samples have low ones. However, Nam et al.~\cite{nam2020learning} use the GCE model as the biased model, which overfit to bias-conflicting samples at the end of the training (Sec.\ref{sec:2.2}), thus bias-conflicting samples have the same low cross-entropy loss as bias-aligned ones. 

Our biased model has no such problem because we protect the bias-conflicting samples from being learned by reducing their weights with the biased predictions. In other words, bias-conflicting samples have
lower weights compared to bias-aligned ones. Therefore, we can naturally use these weights directly for the training of the target model. Specifically, we simply use the inverse of the 
sample weights for biased model as the ones of the target model:
\begin{equation}
    \label{equ:6}
     w(x) = \frac{1}{w_{biased}(x)}
\end{equation}
thus the bias-conflicting samples have high re-weighting values.
\begin{algorithm}[!thb]
    \caption{Echoes}
    \label{alg:Echoes}
    \begin{algorithmic}[1] 
        \STATE \textbf{Input}: dataset $D=\{(x_i,y_i)\}$, biased model $f_\theta$, target model $g_\Theta$, number of iterations $T$, batch size $B$, biased sample weights $W_{B}$, debiased sample weights $W_{D}$, hyperparameter $\lambda$.
        \STATE \textbf{Output}: a biased model $f_\theta$ and a target model $g_\Theta$
        \STATE Initialize $\theta$, $\Theta$, and initialize $W_{B}=\mathbb{1}, W_{D}=\mathbb{0}$
        \STATE \textbf{for} t = 1,\dots,T \textbf{do}
        \STATE \qquad \textbf{for} b = 1,\dots,B \textbf{do}
        \STATE \qquad \qquad Draw a batch $\mathcal{B}=\{(x_i, y_i)\}_{i=1}^{n}$ from $D$
        \STATE \qquad \qquad $\mathcal{L} = W_B\mathcal{L}_{CE}(f_\theta) + \lambda W_D\mathcal{L}_{CE}(g_\Theta)$
        \STATE \qquad \qquad Update $\theta$ and $\Theta$ with $\nabla \mathcal{L}$
        \STATE \qquad \qquad Update $W_B$ (see Alg.~\ref{alg:biased-model})
        \STATE \qquad \qquad Assign $W_{D} = \frac{1}{W_B}$
        \STATE \qquad \qquad Class-level balancing on $W_{D}$ (see Eq.\ref{eq:7})
        \STATE \qquad \textbf{end for}
        \STATE \textbf{end for}
    \end{algorithmic}
\end{algorithm}

~\\
\noindent\textbf{Balancing class weights.} After re-weighting the samples using the above formula, the sum of weights for different classes of samples is likely to be different, resulting in class imbalance. To avoid this problem, we perform class-level weights balancing after applying Eq.\ref{equ:6}. Speciﬁcally, for a sample $(x,y)$ that $y=c\in C$, the weight adjustment is as follows:
\begin{equation}
    \label{eq:7}
    w(x) \coloneqq \frac{ \prod_{g\in C}( \sum_{j\in I_g} w_j )}{ \sum_{i\in I_c} w_i} w(x)
\end{equation}
where $I_c$ is the index set of samples that the label $y=c$. After the adjustment, each class has the same sum of sample weights. The debiased loss for the target model is as follows:
\begin{equation}
    L_{debiased} = W_{debiased}CE(P(X;\theta), Y).
\end{equation}

\begin{table*}[!thb]
    \caption{Debiasing results on CelebA dataset. For classification performance, we report average group accuracy and worst group accuracy. For model fairness, we report gender gap, age gap, and the average of them. We blod the best results for each column.}
    \centering
    \begin{tabular}{c c c | c c | c c c}
    \toprule
        \multirow{2}{*}{Task} & \multirow{2}{*}{Method} & \multirow{2}{*}{bias label} & \multicolumn{2}{c}{Accuracy} & \multicolumn{3}{c}{Bias reliance} \\ \cline{4-5} \cline{6-8}
        & & & Avg group acc & Worst group acc & Gender gap$\downarrow$ & Age gap$\downarrow$ & Avg bias gap$\downarrow$ \\ \midrule
        
        \multirow{7}{*}{Smiling}
        & Vanilla & $\times$ & 76.0\footnotesize{$\pm$0.2} & 36.0\footnotesize{$\pm$1.7} & 41.0\footnotesize{$\pm$0.6} & 11.1\footnotesize{$\pm$1.0} & 26.1\footnotesize{$\pm$0.2}  \\
        & GroupDRO & \checkmark & 76.2\footnotesize{$\pm$0.5} & 37.9\footnotesize{$\pm$1.6} & 40.6\footnotesize{$\pm$1.3} & 11.3\footnotesize{$\pm$0.9} & 25.9\footnotesize{$\pm$1.0}  \\
        & LfF & $\times$ & 75.0\footnotesize{$\pm$0.2} & 35.5\footnotesize{$\pm$2.0} & 42.1\footnotesize{$\pm$0.2} & 9.8\footnotesize{$\pm$1.7} & 25.9\footnotesize{$\pm$0.9}  \\
        & JTT & $\times$ & \textbf{77.8\footnotesize{$\pm$0.6}} & 38.7\footnotesize{$\pm$2.4} & 38.1\footnotesize{$\pm$0.5} & 11.1\footnotesize{$\pm$1.7} & 24.6\footnotesize{$\pm$0.7}  \\
        & DebiAN & $\times$ & 74.5\footnotesize{$\pm$0.6} & 41.1\footnotesize{$\pm$4.3} & 38.0\footnotesize{$\pm$1.4} & 11.7\footnotesize{$\pm$1.3} & 24.9\footnotesize{$\pm$1.4}  \\
        & BPA & $\times$ & 76.3\footnotesize{$\pm$0.6} & 37.1\footnotesize{$\pm$3.8} & 37.6\footnotesize{$\pm$1.6} & 12.1\footnotesize{$\pm$0.9} & 24.9\footnotesize{$\pm$1.2}  \\
        \rowcolor{blue!5} & Echoes & $\times$ & 69.5\footnotesize{$\pm$2.7} & \textbf{58.7\footnotesize{$\pm$4.6}} & \textbf{7.3\footnotesize{$\pm$1.9}} & \textbf{6.8\footnotesize{$\pm$1.8}} & \textbf{7.1\footnotesize{$\pm$1.0}} \\ \midrule
        
        \multirow{7}{*}{\makecell[c]{Narrow\\Eyes}}
        & Vanilla & $\times$ & 60.6\footnotesize{$\pm$0.9} & 13.1\footnotesize{$\pm$0.9} & 62.7\footnotesize{$\pm$1.0} & 19.4\footnotesize{$\pm$0.3} & 41.1\footnotesize{$\pm$0.6} \\
        & GroupDRO & \checkmark & 59.8\footnotesize{$\pm$0.7} & 13.1\footnotesize{$\pm$0.9} & 63.1\footnotesize{$\pm$1.5} & 19.3\footnotesize{$\pm$1.6} & 41.2\footnotesize{$\pm$0.8} \\
        & LfF & $\times$ & 61.2\footnotesize{$\pm$0.4} & 14.7\footnotesize{$\pm$0.7} & 56.9\footnotesize{$\pm$0.6} & 20.9\footnotesize{$\pm$0.2} & 38.9\footnotesize{$\pm$0.2} \\
        & JTT & $\times$ & 61.5\footnotesize{$\pm$0.5} & 12.5\footnotesize{$\pm$1.3} & 60.1\footnotesize{$\pm$1.3} & 21.9\footnotesize{$\pm$0.3} & 41.0\footnotesize{$\pm$0.6} \\
        & DebiAN & $\times$ & \textbf{62.4\footnotesize{$\pm$0.9}} & 16.3\footnotesize{$\pm$1.2} & 56.1\footnotesize{$\pm$1.3} & 19.5\footnotesize{$\pm$0.8} & 37.8\footnotesize{$\pm$0.9} \\
        & BPA & $\times$ & 61.6\footnotesize{$\pm$0.9} & 15.5\footnotesize{$\pm$3.4} & 57.5\footnotesize{$\pm$2.0} & 18.7\footnotesize{$\pm$1.4} & 38.1\footnotesize{$\pm$1.1} \\
        \rowcolor{blue!5} & Echoes & $\times$ & 62.1\footnotesize{$\pm$0.3} & \textbf{48.3\footnotesize{$\pm$3.1}} & \textbf{7.9\footnotesize{$\pm$4.0}} & \textbf{15.7\footnotesize{$\pm$1.8}} & \textbf{11.8\footnotesize{$\pm$2.1}} \\ \midrule

        \multirow{7}{*}{\makecell[c]{Arched\\Eyebrows}}
        & Vanilla & $\times$ & 61.9\footnotesize{$\pm$0.4} & 9.6\footnotesize{$\pm$1.5} & 63.5\footnotesize{$\pm$0.5} & 19.9\footnotesize{$\pm$0.4} & 41.7\footnotesize{$\pm$0.4} \\
        & GroupDRO & \checkmark & 60.4\footnotesize{$\pm$0.5} & 12.3\footnotesize{$\pm$3.1} & 62.2\footnotesize{$\pm$1.0} & \textbf{18.2\footnotesize{$\pm$1.3}} & 40.2\footnotesize{$\pm$0.2} \\
        & LfF & $\times$ & 60.8\footnotesize{$\pm$0.4} & 14.9\footnotesize{$\pm$1.7} & 56.9\footnotesize{$\pm$1.0} & 22.5\footnotesize{$\pm$0.6} & 39.7\footnotesize{$\pm$0.6} \\
        & JTT & $\times$ & \textbf{62.4\footnotesize{$\pm$0.3}} & 12.8\footnotesize{$\pm$0.6} & 59.5\footnotesize{$\pm$0.9} & 21.8\footnotesize{$\pm$0.6} & 40.7\footnotesize{$\pm$0.7} \\
        & DebiAN & $\times$ & 61.3\footnotesize{$\pm$0.3} & 13.9\footnotesize{$\pm$0.9} & 58.3\footnotesize{$\pm$1.3} & 18.7\footnotesize{$\pm$1.2} & 38.5\footnotesize{$\pm$0.3} \\
        & BPA & $\times$ & 61.7\footnotesize{$\pm$0.1} & 13.2\footnotesize{$\pm$1.0} & 58.1\footnotesize{$\pm$0.6} & 22.9\footnotesize{$\pm$1.4} & 40.5\footnotesize{$\pm$0.4} \\
        \rowcolor{blue!5} & Echoes & $\times$ & 58.0\footnotesize{$\pm$1.6} & \textbf{35.5\footnotesize{$\pm$2.9}} & \textbf{7.5\footnotesize{$\pm$3.1}} & 21.7\footnotesize{$\pm$3.0} & \textbf{14.6\footnotesize{$\pm$0.5}} \\
    \bottomrule
    \end{tabular}
    \label{tab:main-res-celeba}
\end{table*}

\subsection{Complete Method}
\label{sec:3.3}

We now have an approach for training a biased model and an approach for training a target model once we have a biased model. We follow the training strategy of Nam et al.~\cite{nam2020learning} that jointly train the bias model and the target model using followed loss:
\begin{equation}
    \mathcal{L} = \mathcal{L}_{biased} + \lambda \mathcal{L}_{debiased}
\end{equation}
where $\lambda$ is a hyperparameter for balancing the loss.
Fig.\ref{fig:model} illustrates the complete method, and the pseudo-code of \textit{Echoes} is presented in Alg.\ref{alg:Echoes}.

\section{Experiment}

We now validate the effectiveness of the proposed method on debiasing benchmarks. We first introduce the experimental settings in Section~\ref{sec:4.1}, then, we introduce the experimental results in Section~\ref{sec:4.2}, \ref{sec:4.3} and \ref{sec:4.4}.

\subsection{Setup}
\label{sec:4.1}

\begin{figure}[!thb]
\setlength{\abovecaptionskip}{0.2em} 
    \centering
    \includegraphics[width=0.8\linewidth]{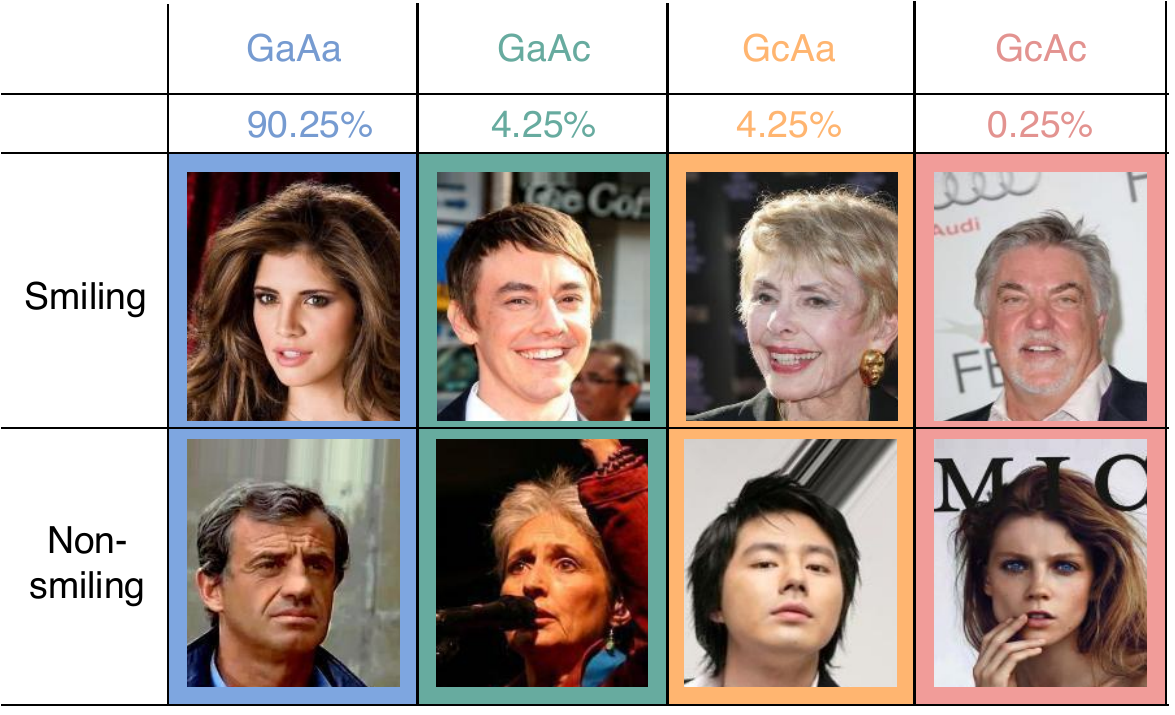}
    \caption{Unbalanced groups in CelebA’s training set based on two biased attributes: \textit{Gender} and \textit{Age}. And \textit{GaAc} means samples that are gender-aligned and age-conflicting.}
    \label{fig:celeba}
    \vspace{-0.2cm}
\end{figure}

\noindent\textbf{Datasets.} \\
\textbf{CelebA}~\cite{liu2015faceattributes} is a large-scale face dataset for face image recognition, containing 40 attributes for each image, of which \textit{Gender} and \textit{Age} are protected attributes that should be prevented from being learned by the model. However, most existing works design and evaluate methods under the assumption that a single bias is present in the data. We set both \textit{gender} and \textit{age} as the biased attributes, and set \textit{Smiling}, \textit{Narrow Eyes} and \textit{Arched Eyebrows} as the target attributes. We set the skew ratio to 95\% for both \textit{gender} and \textit{age}, and sample 8000 images in the original CelebA as the training set. The data can be divided into 8 groups using two bias attributes and one target attribute. Fig.\ref{fig:celeba} shows the training data distribution for the \textit{Smiling} classification task. For the test set, we sample 1000 images, and each group has an equal number of samples (i.e., 125).

\begin{table*}[!thb]
    \caption{Debiasing results on UrbanCars dataset. For model fairness, we report background (BG) gap, co-occurring object (Co-obj) gap, and the average of BG gap and Co-obj gap.}
    \centering
    \begin{tabular}{c c c | c c | c c c}
    \toprule
        \multirow{2}{*}{Task} & \multirow{2}{*}{Method} & \multirow{2}{*}{bias label} & \multicolumn{2}{c}{Accuracy} & \multicolumn{3}{c}{Bias reliance} \\ \cline{4-5} \cline{6-8}
        & & & Avg group acc & Worst group acc & BG gap$\downarrow$ & Co-obj gap$\downarrow$ & Avg bias gap$\downarrow$ \\ \midrule
        \multirow{8}{*}{\makecell[c]{Car\\Object}}
        & Vanilla & $\times$ & 60.6\footnotesize{$\pm$0.3} & 19.2\footnotesize{$\pm$1.5} & 48.9\footnotesize{$\pm$1.0} & 23.7\footnotesize{$\pm$0.5} & 36.3\footnotesize{$\pm$0.4}  \\
        & GroupDRO & \checkmark & 60.7\footnotesize{$\pm$0.7} & 20.8\footnotesize{$\pm$1.1} & 46.8\footnotesize{$\pm$2.3} & 20.4\footnotesize{$\pm$0.8} & 33.6\footnotesize{$\pm$0.7}  \\
        & LfF & $\times$ & 64.0\footnotesize{$\pm$1.5} & 35.7\footnotesize{$\pm$5.4} & 27.7\footnotesize{$\pm$3.8} & \textbf{5.7\footnotesize{$\pm$2.8}} & 16.7\footnotesize{$\pm$1.8}  \\
        & JTT & $\times$ & \textbf{69.5\footnotesize{$\pm$2.3}} & 33.3\footnotesize{$\pm$6.9} & 39.3\footnotesize{$\pm$3.1} & 13.9\footnotesize{$\pm$0.7} & 26.6\footnotesize{$\pm$1.8}  \\
        & DebiAN & $\times$ & 62.4\footnotesize{$\pm$0.9} & 27.2\footnotesize{$\pm$2.5} & 38.1\footnotesize{$\pm$2.5} & 15.0\footnotesize{$\pm$1.5} & 26.5\footnotesize{$\pm$1.9}  \\
        & BPA & $\times$ & 63.8\footnotesize{$\pm$0.8} & 21.6\footnotesize{$\pm$4.9} & 32.3\footnotesize{$\pm$0.8} & 29.7\footnotesize{$\pm$2.1} & 31.0\footnotesize{$\pm$0.7}  \\
        \rowcolor{blue!5} & Echoes & $\times$ & 63.2\footnotesize{$\pm$0.7} & \textbf{48.5\footnotesize{$\pm$2.8}} & \textbf{4.8\footnotesize{$\pm$2.2}} & 15.7\footnotesize{$\pm$0.3} & \textbf{10.3\footnotesize{$\pm$1.2}} \\
    \bottomrule
    \end{tabular}
    \label{tab:main-res-urbancars}
\end{table*}

\begin{table}[!thb]
    \caption{Debiasing results on BFFHQ dataset. \textit{Ratio} represents the ratio of bias-conflicting samples in the training set.}
    \centering
    \resizebox{.99\linewidth}{!}{
    \begin{tabular}{c c | c c c c c c}
    \toprule
        \makecell{Ratio(\%)} & Metrics & Vanilla & LfF & JTT & DebiAN & BPA & Echoes  \\ \midrule
        \multirow{3}{*}{0.5}   & Avg & 72.8 & 73.4 & \textbf{76.0} & 74.7 & 74.7 & 71.7 \\
                                & Worst & 35.6 & 33.2 & 40.8 & 40.0 & 37.2 & \textbf{48.8} \\
                                & Bias & 52.6 & 50.8 & 47.0 & 49.4 & 47.8 & \textbf{38.2} \\ \midrule
        \multirow{3}{*}{1.0}   & Avg & 74.4 & 73.5 & \textbf{76.2} & 74.5 & 74.9 & 70.9 \\
                                & Worst & 40.4 & 30.0 & 41.2 & 39.6 & 45.2 & \textbf{52.0} \\
                                & Bias & 41.6 & 50.6 & 45.2 & 47.4 & 44.2 & \textbf{19.8} \\ \midrule
        \multirow{3}{*}{2.0}   & Avg & 78.5 & 77.2 & \textbf{80.4} & 78.2 & 79.9 & 73.5 \\
                                & Worst & 50.0 & 48.4 & 55.6 & 55.6 & 54.8 & \textbf{60.0} \\
                                & Bias & 40.2 & 40.4 & 36.8 & 38.4 & 37.4 & \textbf{21.0} \\
        
    \bottomrule
    \end{tabular}
    }
    \label{tab:main-res-bffhq}
\end{table}

\begin{figure}
\setlength{\abovecaptionskip}{0em} 
\setlength{\belowcaptionskip}{-1em} 
    \centering
    \subfigure[CelebA]{
        \includegraphics[height=81px]{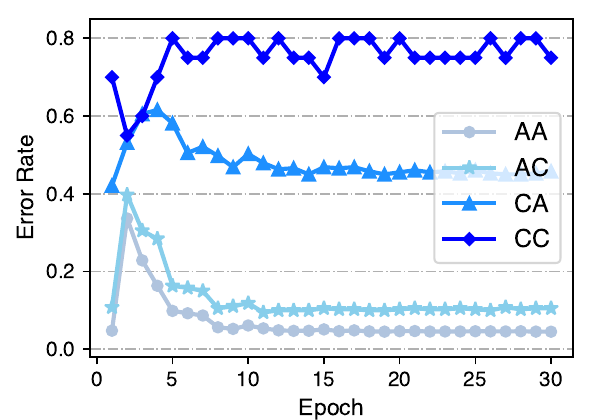}
    }
    \subfigure[UrbanCars]{
        \includegraphics[height=81px]{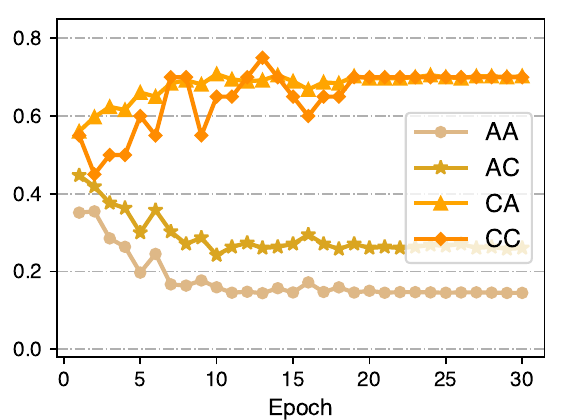}
    }
    \caption{The classiﬁcation error rate of our biased model on the training data of CelebA and UrbanCars.}
    \label{fig:train-error-Echoes}
\end{figure}

\noindent\textbf{UrbanCars} is a synthetic car dataset created by Li et al.~\cite{li_2023_whac_a_mole} which has two biased attributes: \textit{background (bg)} and \textit{co-occurring object (co-obj)}. Li et al.~\cite{li_2023_whac_a_mole} segment and recombine the images of the three datasets: Stanford Cars~\cite{krause20133d} for car images, Places~\cite{zhou2017places} for background images, and LVIS~\cite{gupta2019lvis} for co-occurring object images. The cars, the backgrounds, and the co-occurring objects can all be divided into $\{urban, country\}$. The skew ratios of the biased attributes in the train set and the test set are the same as the CelebA dataset, i.e., 8000 unbalanced images for training and 1000 balanced images for testing.

\noindent\textbf{BFFHQ} (Biased FFHQ)~\cite{lee2021learning} is constructed based on the real-world face dataset FFHQ~\cite{karras2019style}, where the label is age and the biased feature is gender.

~\\
\noindent\textbf{Baselines.} To evaluate the effectiveness of our method in debiasing, we compare it with the prior methods including vanilla network, GroupDRO~\cite{Sagawa2020Distributionally}, LfF~\cite{nam2020learning}, JTT~\cite{liu2021just}, DebiAN~\cite{li2022discover}, BPA~\cite{seo2022unsupervised}. Vanilla denotes the classiﬁcation model trained only with the original cross-entropy (CE) loss, without any debiasing strategies. GroupDRO explicitly leverages the bias labels (e.g., the gender labels in CelebA) during the training phase, while others require no prior knowledge of the biases. 
LfF and JTT are evaluated on the single-bias datasets only, while DebiAN and BPA are evaluated on the multi-bias datasets in their respective works.
\begin{table}[!th]
    \caption{The bias-aligned accuracy and bias-conflicting accuracy on the test sets of CelebA and UrbanCars.}
    \centering
    \begin{tabular}{cc|ccc}
    \toprule
        Dataset & Accuracy & ERM & GCE & Ours \\ \midrule
        \multirow{2}{*}{CelebA} & Bias-aligned$\uparrow$ & 98.3 & 98.1 & \textbf{98.4} \\
                                & Bias-conflicting$\downarrow$ & 68.5 & 68.4 & \textbf{65.5} \\ \midrule
        \multirow{2}{*}{UrbanCars} & Bias-aligned$\uparrow$ & \textbf{93.5} & 91.7 & 91.9 \\
                                   & Bias-conflicting$\downarrow$ & 49.6 & 49.9 & \textbf{45.3} \\        
        
    \bottomrule
    \end{tabular}
    \label{tab:test-result-biased-model}
    \vspace{-1em}
\end{table}

\noindent\textbf{Points of comparison.} We aim to answer two main questions: (1) How does the classification performance of \textit{Echoes} compare with other methods? (2) How does the fairness of \textit{Echoes} compare to other methods? To answer the first question, we report the average group accuracy and the worst group accuracy. 
To answer the second question, we check whether the model predictions meet the fairness criterion \textit{bias gap}, which indicates the drop in model accuracy when the group shifts. For example, the \textit{gender gap} in CelebA is as follows:
\begin{equation}
    \begin{aligned}
    gender\;gap&=\frac{|ACC_{GaAa}-ACC_{GcAa}| + |ACC_{GaAc}-ACC_{GcAc}|}{2} \\
    \end{aligned}
\end{equation}

\noindent\textbf{Implementation details.} In the experiments, we use ResNet-18~\cite{he2016deep} for all the datasets. For the training, we set the batch size as 256 and the learning rate as 3e-4. We train the models for 100 epochs with the Adam~\cite{DBLP:journals/corr/KingmaB14} optimizers in the experiments. We set $t_{error}=0.3$ for UrbanCars, $t_{error}=0.5$ for CelebA and BFFHQ, and $\alpha=0.5, \lambda=1000$ for all datasets.
Since there is no prior knowledge of the biases for model selection, we report the accuracy of the last epoch. All experimental results have averaged over three independent trials.

\subsection{Main Result}
\label{sec:4.2}
\begin{table}[!th]
    \caption{The comparison of the results of our method with and without class balancing on CelebA.}
    \centering
    \begin{tabular}{cc|ccc}
    \toprule
        Task & Metrics & \makecell{w/o class\\balancing} & \makecell{w/ class\\balancing} \\ \midrule
        \multirow{2}{*}{Smiling} & Avg & 67.5 & \textbf{69.5} \\
                                & Worst & 54.1 & \textbf{58.7} \\ \midrule
        \multirow{2}{*}{\makecell{Narrow\\Eyes}} & Avg & 60.3 & \textbf{62.1} \\
                                & Worst & 41.0 & \textbf{48.3} \\ \midrule
        \multirow{2}{*}{\makecell{Arched\\Eyebrows}} & Avg & 56.7 & \textbf{58.0} \\
                                & Worst & 35.2 & \textbf{35.5} \\
    \bottomrule
    \end{tabular}
    \label{tab:class-balance}
\end{table}

\noindent\textbf{Debiasing with multi-bias.} Table~\ref{tab:main-res-celeba} and Table~\ref{tab:main-res-urbancars} show the quantitative experimental results on CelebA and UrbanCars, respectively. On CelebA, our method achieves the best worst group accuracy and average bias gap on all three tasks. The average group accuracy of our method is relatively low, but the worst group accuracy is high, which indicates that the accuracy of different groups is close, and this also reflects the debiasing performance of our method. On UrbanCars, our method also achieves the best worst group accuracy and average bias gap which shows the effectiveness of our method in handling multi-bias.

\noindent\textbf{Debiasing with single-bias.} Table~\ref{tab:main-res-bffhq} shows the results on BFFHQ~\cite{lee2021learning}. The bias is the accuracy drop from 
bias-aligned accuracy to bias-conflicting accuracy. The results show that our method also outperforms baselines in debiasing under single-bias setting.


\subsection{The Effectiveness of Our Biased Model}
\label{sec:4.3}
For the biased model, we want it to (i) not overfit to the bias-conflicting samples in the training set, and (ii) fully learn the biased features in order to pass information to the target model. Fig.\ref{fig:train-error-Echoes} and Table~\ref{tab:test-result-biased-model} show the advantages of our bias model compared to ERM model and GCE model. 

As show in Fig.\ref{fig:train-error-Echoes}, the error rate of our model on the bias-conflicting samples of the training set is consistently high, i.e., our biased model does not overfit to the bias-conflicting samples. As show in Table~\ref{tab:test-result-biased-model}, our biased model achieves the lowest bias-conflicting accuracy on both test sets of CelebA and Urbancars and the highest bias-aligned accuracy on CelebA, which indicates that the model sufficiently learns the biased features rather than the target features.

\subsection{Ablation Study}
\label{sec:4.4}
\noindent\textbf{Class-level balancing.} Table~\ref{tab:class-balance} shows the comparison of the results of \textit{Echoes} with and
without class-level balancing on CelebA. Both the average gourp and worst group accuracy of the model decrease on all tasks when without class-level balancing, indicating the effectiveness of the class-level balancing.

\begin{figure}
\setlength{\abovecaptionskip}{0em} 
\setlength{\belowcaptionskip}{-2em} 
    \centering
    \includegraphics[width=0.95\linewidth]{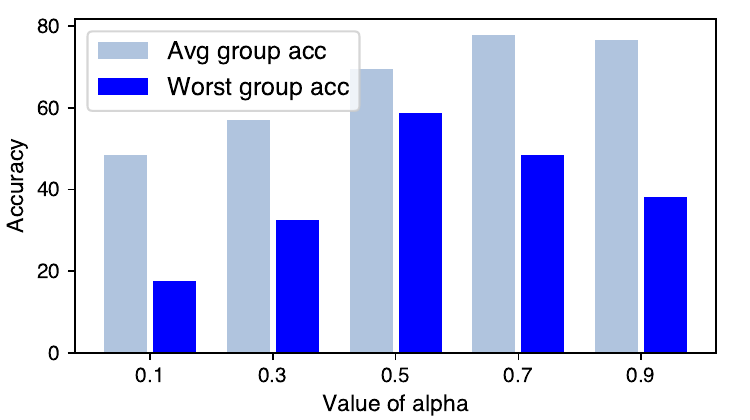}
    \caption{Accuracy of \textit{Smiling} classification task with different values of alpha $\alpha$ on CelebA.}
    \label{fig:down_weight}
\end{figure}

~\\
\noindent\textbf{Different values of $\alpha$.} We use $\alpha\in[0,1]$ to control the degree of weight reduction of the samples when training the biased model. Fig.\ref{fig:down_weight} shows the results of different values of $\alpha$. When $\alpha$ is close to 0, the model ignores too many samples, resulting in insufficient learning, so both worst group and average group accuracy are low. When $\alpha$ is close to 1, the model is similar to ERM, leading to low worst group accuracy. Therefore, taking $\alpha$ as 0.5 is appropriate.

\section{Related work}

\noindent\textbf{Debiasing methods.} Early works on debiasing assume some prior knowledge about the bias (i.e., supervised debiasing). Some methods require that each training sample is provided with a bias label (e.g., the backgrouds of the car images)~\cite{Sagawa2020Distributionally, tartaglione2021end, hong2021unbiased}. Other methods use knowledge of the bias type in the datasets, such as color or texture in images, and this information is typically used to design custom architectures~\cite{bahng2020learning, GeirhosRMBWB19}. For example, Geirhos et al.~\cite{GeirhosRMBWB19} use a style transfer network to generate StylizedImageNet, a stylized version of ImageNet. Models trained on such data can mitigate texture bias. Rencet works focus on the more realistic and challenging setting, where debiasing without bias information (i.e., unpervised debiasing)~\cite{nam2020learning, liu2021just, lee2021learning}. The works assume that biased features are "easy-to-learned" than robust ones and thus train an auxiliary model that intentionally relies primarily on the biased features. However, the existing proposed auxiliary models (e.g., ERM model~\cite{liu2021just} or GCE model~\cite{nam2020learning}) overfit to the bias-conflicting samples of the train data which impairs the debiasing performance of the method. In contrast, our method constructes an echo chamber environment (i.e., reduces the weights of bias-conflicting samples) when training the auxiliary model, which mitigates the overfitting.

~\\
\noindent\textbf{Echo chamber effect.} The echo chamber effect is a common phenomenon in social media and recommender systems~\cite{ge2020understanding, nguyen2014exploring}. Participants are exposed to homogenized views, which are repeatedly disseminated and diffused, eventually leading to the polarization of information and the formation of a closed system. For example, users in the recommendation system will constantly click on the items of interest and ignore the uninteresting ones, the recommendation system will optimize the recommendation results, and finally the items displayed will be of interest to the users. Although this can introduce exposure bias~\cite{khenissi2020modeling} in the recommender system, we use the echo chamber effect to amplify the bias in the training data so that the biased model can better learn biased features.

\section{Conclution}

We presented \textit{Echoes}, a method for unsupervised debiasing that jointly trains a biased model and a target model in an echo chamber environment. The echo chamber is constructed by continuously reducing the weights of samples misclassified by the biased model under the "easy-to-learned" hypothesis. Experiments show that \textit{Echoes} outperforms baselines both on multi-biased and single-biased benchmarks.

\begin{acks}
This work is supported by the Fundamental Research Funds for the Central Universities (No. 2023JBZY033), the National Natural Science Foundation of China (No. 61832002, 62172094), the Beijing Natural Science Foundation (No. JQ20023), and CCF-Zhipu AI Large Model Foundation.
\end{acks}

\bibliographystyle{ACM-Reference-Format}
\balance
\bibliography{sample-base}

\end{document}